\documentclass[10pt]{article}
\usepackage[utf8]{inputenc}
\usepackage{amsmath,amssymb,amsthm}
\usepackage{geometry}
\usepackage{hyperref}
\usepackage{multirow}
\usepackage{tabularx}

\theoremstyle{definition}

\usepackage{graphicx} 
\usepackage{booktabs}
\usepackage{subcaption}
\usepackage{algorithm}
\usepackage{algorithmic}
\usepackage{rotating}
\usepackage{float}
\usepackage{array}
\usepackage{makecell}
\usepackage{placeins}

\title{Interpretable Tissue Architecture Complements Pathology Foundation Models}

\author{
	Zeeshan Shahid$^{1,*}$,
	Hammam M. AlGhamdi$^{2,*}$\\[0.6em]
	{\small $^{1}$Department of Mathematics,}\\
	{\small COMSATS University Islamabad, Pakistan}\\[0.4em]
	{\small $^{2}$Department of Computer Science \& Artificial Intelligence,}\\
	{\small College of Computer Science and Engineering,}\\
	{\small University of Jeddah, 21493, Saudi Arabia}\\[0.6em]
	{\small $^{*}$Corresponding authors}\\[0.4em]
	{\small Email: zeeshanshahid9515@gmail.com (Z. Shahid)}\\
	{\small Email: hmsalghamdi@uj.edu.sa (H. AlGhamdi)}
}

\begin{document}
	
	\maketitle
	

\begin{abstract}
Tissue architecture provides important information for histopathological assessment, yet its complementary predictive value alongside pathology foundation-model representations remains insufficiently understood. This study investigates whether interpretable measurements of cellular and multicellular tissue organization provide additional predictive information beyond learned representations. We introduce a 139-dimensional cellular-to-community representation comprising nine feature groups characterizing nuclear morphology, cellular composition, spatial connectivity, community organization, and cell–cell interactions. Their predictive contributions are systematically evaluated independently and through feature-level fusion with CTransPath and PLIP embeddings using multiple-instance learning (MIL). On the BCNB dataset, estrogen receptor (ER) and progesterone receptor (PR) classification is evaluated across four MIL architectures using five-fold patient-disjoint cross-validation. Selected CTransPath fusion configurations achieved mean ROC-AUC values of 0.9144 (ER) and 0.8360 (PR), while PLIP fusion achieved 0.8791 and 0.8015, respectively. Patient-level analyses identified ER-associated differences in nuclear morphology, graph connectivity, cellular communities, and cell–cell interactions. On LungHist700, selected fusion configurations achieved higher mean ROC-AUC across four classification tasks, including adenocarcinoma versus squamous cell carcinoma (0.7967 versus 0.7137 for CTransPath alone). Overall, these findings suggest that interpretable tissue-architecture descriptors can complement foundation-model representations while providing measurable insights into histopathological organization.
\end{abstract}


\section{Introduction}

Histopathological assessment relies on cellular morphology,
tissue organization, and spatial relationships among different
cell populations to characterize disease-associated changes.
Advances in computational pathology have enabled quantitative
analysis of hematoxylin and eosin (H\&E)-stained whole-slide
images (WSIs), supporting automated tissue classification and
biomarker prediction. However, systematically characterizing
tissue architecture across cellular and multicellular scales
and determining whether explicit architectural measurements
provide predictive information complementary to learned
visual representations remain important research challenges.

Breast cancer is the most frequently diagnosed cancer among
women worldwide and remains a leading cause of cancer-related
mortality \cite{bray2024globocan}. Accurate assessment of
clinically important biomarkers, including estrogen receptor
(ER), progesterone receptor (PR), and human epidermal growth
factor receptor 2 (HER2), is essential for molecular subtyping,
prognosis, and treatment selection \cite{WHO2020Breast}.
These biomarkers are routinely assessed using
immunohistochemistry (IHC) and, where appropriate, additional
molecular testing. Nevertheless, H\&E-stained WSIs contain
morphological and spatial information associated with tumor
biology. Quantifying histological characteristics associated
with biomarker status may therefore provide complementary
information for computational pathology and help characterize
relationships between tissue architecture and molecular
phenotype.

Previous studies have demonstrated the potential of predicting
breast cancer biomarkers directly from H\&E-stained images.
Associations between nuclear morphological characteristics
and hormone receptor status suggest that molecular phenotypes
may be reflected in measurable histopathological patterns
\cite{li2021nuclear}. Deep learning approaches have further
enabled the extraction of discriminative visual representations
for biomarker prediction \cite{gamble2021determining}.
Multiple instance learning (MIL) has become particularly
important for WSI analysis because it aggregates patch-level
information into slide-level predictions without requiring
individual patch annotations. Attention-based MIL additionally
allows different tissue regions to contribute unequally to
the prediction \cite{ilse2018attention}. Graph-based methods
have also demonstrated the predictive utility of spatial
tissue structure; for example, SlideGraph+ employed graph
representations for HER2 prediction from H\&E WSIs
\cite{lu2022slidegraph+}.

Beyond individual nuclear characteristics, tissue architecture
depends on the spatial arrangement and interactions of
different cell populations. Cellular graphs represent these
relationships by modeling detected cells as vertices and
their spatial proximity as edges. Javed \emph{et al.}
\cite{javed2020cellular} demonstrated that cellular communities
derived from such graphs can characterize tissue phenotypes
and higher-order spatial organization. Other investigations
have emphasized the importance of spatial relationships
among tumor, stromal, and immune cell populations in
characterizing tissue heterogeneity
\cite{ali2024spatial,chen2024spatial}. Together, these
studies indicate that cellular composition, nuclear
morphology, graph connectivity, community organization,
and cell--cell interactions describe related but distinct
aspects of tissue architecture. Their predictive contributions,
however, may depend on the classification task and the
combination of architectural characteristics considered.

In parallel, pathology foundation models have enabled
transferable visual representations learned from large
collections of histopathological images. CTransPath combines
convolutional and multi-scale Transformer components to
capture local and global visual information
\cite{wang2022ctranspath}, while Pathology Language and Image
Pre-Training (PLIP) learns pathology-specific representations
through image--text pretraining \cite{huang2023plip}.
More recently, ALR-Mamba introduced a hierarchical
long-range representation framework for hormone receptor
prediction from breast cancer WSIs, including evaluation
on the BCNB dataset \cite{liu2026alr}. Although learned
representations capture complex morphological and contextual
patterns, their embedding dimensions do not directly
correspond to predefined measurements of nuclear morphology,
cellular connectivity, or multicellular organization.
Conversely, explicit tissue-architecture descriptors provide
measurable and interpretable characteristics of histological
organization but may not capture the full visual information
encoded by foundation-model embeddings.

These complementary perspectives motivate an important
research question: to what extent do explicit measurements
of cellular and multicellular tissue architecture provide
predictive information independently of, and in combination
with, pathology foundation-model representations?
Addressing this question requires examining not only the
standalone predictive utility of architectural descriptors
but also how their combinations contribute when integrated
with learned visual representations. Importantly, strong
standalone classification performance does not necessarily
imply that a representation will provide additional
predictive value when combined with another representation.
A systematic evaluation is therefore needed to investigate
the task-dependent contributions of interpretable tissue
architecture and its potential complementarity with
foundation-model embeddings.

To investigate these questions, we introduce an interpretable
cellular-to-community representation comprising 139 descriptors
organized into nine feature groups. These descriptors
characterize cellular composition, nuclear morphology,
spatial graph structure, community organization, centrality,
spectral properties, and cell--cell interactions. We
systematically evaluate all $2^9-1=511$ nonempty feature-group
combinations for ER and PR classification on BCNB using
five-fold patient-disjoint cross-validation. Their
complementary predictive value is investigated through
independent feature-level fusion with CTransPath and PLIP
embeddings across four MIL architectures: Gated Attention MIL,
ALR-Mamba, CLAM-SB, and CLAM-MB
\cite{ilse2018attention,lu2021data}. We additionally examine
patient-level associations between interpretable descriptors
and ER status to identify measurable tissue characteristics
associated with hormone receptor expression. Finally, we
evaluate the representation on LungHist700 across four
pulmonary histopathology classification tasks involving
adenocarcinoma (ACA), squamous cell carcinoma (SCC), and
normal tissue to investigate its applicability beyond
breast cancer.

The main contributions of this work are summarized as follows:

\begin{itemize}
	
	\item We introduce a 139-dimensional interpretable
	cellular-to-community representation integrating
	nuclear morphology, cellular composition, spatial
	graph properties, community organization, and
	cell--cell interactions into nine feature groups
	that characterize multiple levels of tissue
	architecture.
	
	\item We systematically investigate all
	$2^9-1=511$ nonempty feature-group combinations
	to assess how the composition and number of
	architectural descriptors relate to observed
	ER and PR classification performance using
	five-fold patient-disjoint cross-validation.
	
	\item We investigate the complementary predictive
	value of interpretable tissue-architecture
	descriptors through independent feature-level
	fusion with CTransPath and PLIP embeddings
	across four MIL architectures, comparing
	selected fusion configurations with their
	corresponding foundation-model-only baselines.
	
	\item We identify patient-level associations
	between ER status and measurable nuclear,
	graph-based, community-level, and cell-interaction
	characteristics, and further evaluate the
	representation on LungHist700 to investigate
	its applicability to pulmonary histopathological
	classification.
	
\end{itemize}

\section{Related Work}
Quantitative histopathology has long investigated the
relationship between cellular morphology and breast cancer
phenotypes. Li \emph{et al.}~\cite{li2021nuclear} reported
associations between nuclear morphological characteristics
and ER and HER2 expression, indicating that biomarker-related
information can be reflected in measurable nuclear features.
Nuclear morphology and pleomorphism are also established
components of conventional histopathological assessment
\cite{WHO2020Breast}. Computational approaches have therefore
quantified nuclear area, perimeter, axis lengths, eccentricity,
solidity, and circularity to provide reproducible descriptions
of cellular morphology, including automated assessment of
nuclear pleomorphism \cite{li2021nuclear,teoh2024quantitative}.

Deep learning has expanded biomarker prediction beyond
predefined morphological measurements. Gamble
\emph{et al.}~\cite{gamble2021determining} demonstrated the
prediction of clinically relevant breast cancer biomarkers
from routine H\&E-stained slides and investigated morphological
characteristics associated with model predictions. These
studies establish the relevance of histological morphology
to biomarker prediction. Nevertheless, individual nuclear
measurements do not fully characterize the spatial
relationships and multicellular arrangements that contribute
to tissue architecture.

Graph-based histopathology methods explicitly represent
spatial tissue organization by modeling cells or tissue
regions as nodes and their relationships as edges. Such
representations enable the characterization of connectivity,
cellular neighborhoods, and spatial interactions that cannot
be described by isolated morphological measurements.

SlideGraph+~\cite{lu2022slidegraph+} incorporated graph-based
representations and slide-level aggregation for HER2
prediction, demonstrating the predictive relevance of
spatial tissue context. At a higher organizational scale,
Javed \emph{et al.}~\cite{javed2020cellular} formulated
tissue phenotyping through cellular-community detection,
showing that groups of spatially related cells can
characterize distinct tissue phenotypes. This approach was
subsequently extended to multiplex cellular communities,
which integrate multiple cellular and contextual
representations into graph-based tissue models
\cite{javed2020multiplex}.

These studies demonstrate the importance of representing
tissue organization beyond individual nuclei. However,
morphological, graph-based, and community-level descriptors
characterize different aspects of histological structure.
Their relative predictive contributions, particularly when
considered in different combinations, require systematic
evaluation. Moreover, the extent to which explicitly
measured tissue architecture contributes predictive
information beyond learned image representations remains
an important question.

Multiple instance learning (MIL) has become a widely used
framework for WSI classification because it enables
slide-level prediction from collections of image patches
without requiring patch-level labels. Attention-based MIL
further learns the relative contributions of individual
instances during aggregation \cite{ilse2018attention}.

Pathology foundation models provide transferable visual
representations through large-scale pretraining.
CTransPath combines convolutional and multi-scale
Transformer components with self-supervised learning to
obtain histopathological representations
\cite{wang2022ctranspath}. UNI has demonstrated broad
transferability across computational pathology tasks
\cite{chen2024uni}. For breast cancer biomarker prediction,
ALR-Mamba introduced a hierarchical long-range
representation framework and evaluated hormone receptor
classification on datasets including BCNB
\cite{liu2026alr}.

These approaches demonstrate the effectiveness of learned
visual representations and advanced aggregation mechanisms.
However, their feature dimensions are not explicitly defined
as measurements of nuclear morphology, cell-type composition,
spatial connectivity, or multicellular communities. This
limits direct interpretation of individual embedding
coordinates in terms of predefined histological properties.
Conversely, explicitly constructed tissue descriptors
provide measurable architectural characteristics but may
not capture the full visual information represented by
foundation-model embeddings.

Existing research has established the relevance of nuclear
morphology, spatial tissue organization, cellular communities,
and learned histopathological representations. However,
the reviewed studies do not establish how the predictive
contributions of multiple interpretable architectural
feature groups vary across combinations or how these
descriptors complement pathology foundation-model embeddings
under different MIL aggregation architectures.

The present study addresses this question through a
139-dimensional cellular-to-community representation
comprising nine interpretable feature groups. Rather than
introducing a new MIL architecture, we systematically
investigate the independent and complementary predictive
value of these descriptors through feature-group
combinations and feature-level fusion with CTransPath
and PLIP embeddings. Evaluation across breast cancer
biomarker prediction and pulmonary histopathological
classification examines the applicability of this
representation across different classification tasks,
while patient-level association analyses investigate
measurable tissue characteristics related to ER status.

\section{Materials and Method}

\subsection{Datasets}

Two histopathology datasets were used to evaluate the
proposed framework: the Early Breast Cancer Core-Needle
Biopsy Whole-Slide Image (BCNB) dataset
\cite{xu2021predicting} and the multi-class LungHist700
dataset \cite{diosdado2024lunghist700}.

\noindent\textbf{BCNB dataset.}
The BCNB dataset comprises 1,058 patients with early-stage
breast cancer and corresponding hematoxylin and eosin
(H\&E)-stained primary-tumor core-needle biopsy whole-slide
images (WSIs). Although the dataset includes patient-level
clinicopathological information, only H\&E WSIs and their
corresponding biomarker labels were used in this study.
Estrogen receptor (ER) and progesterone receptor (PR)
status were evaluated as separate binary classification
tasks.

\noindent\textbf{LungHist700 dataset.}
LungHist700 contains 691 high-resolution pulmonary tissue
images from 45 patients \cite{diosdado2024lunghist700}.
The dataset was used to investigate the applicability of
the proposed representation beyond breast cancer.
Four classification tasks were considered:
adenocarcinoma (ACA) versus normal tissue,
squamous cell carcinoma (SCC) versus normal tissue,
ACA versus SCC, and three-class ACA--SCC--normal
classification.

Images acquired at both $20\times$ and $40\times$
magnification were included. Five-fold patient-disjoint
cross-validation was performed using patient identifiers
as grouping variables, ensuring that images from the same
patient, including those acquired at different
magnifications, were assigned to the same fold.

\subsection{Nuclei Segmentation and Patch Preparation}
\label{subsec:nuclei_segmentation}

Histopathological images were processed as non-overlapping
$512\times512$-pixel patches. Nuclear instance segmentation
and cell-type classification were performed using HoVer-Net
\cite{graham2019hovernet}, implemented through TIAToolbox
\cite{pocock2022tiatoolbox}. Each detected nucleus was
represented by its centroid coordinates, nuclear contour,
and predicted cell type.

Five nuclear categories were considered: neoplastic,
inflammatory, connective, dead, and non-neoplastic
epithelial. Before feature extraction, detections with
missing centroid or contour information, invalid cell-type
identifiers, fewer than five contour points, or nonpositive
contour area were excluded. Patches containing fewer than
ten valid nuclei were also excluded.

The retained nuclear centroids, contours, and predicted
cell types were subsequently used to construct the
patch-level cellular-to-community representation.

\subsection{Cellular-to-Community Representation}
\label{subsec:cellular_community}

To quantify tissue architecture at multiple organizational
levels, we construct a 139-dimensional patch-level
representation comprising nine interpretable feature
groups. These descriptors characterize cellular
composition, nuclear morphology, spatial connectivity,
community organization, graph centrality, spectral
properties, and cell--cell interactions.

\noindent\textbf{Spatial graph construction.}
For each eligible patch, an undirected spatial graph
$G=(V,E)$ is constructed, where each vertex $v_i\in V$
represents a detected nucleus with centroid
$\mathbf{x}_i\in\mathbb{R}^2$. Two distinct nuclei are
connected when their Euclidean centroid distance is
smaller than a fixed threshold of $r=40$ pixels:
\begin{equation}
	\{v_i,v_j\}\in E
	\quad\Longleftrightarrow\quad
	0<\|\mathbf{x}_i-\mathbf{x}_j\|_2<r,
	\qquad i<j.
\end{equation}
The original pixel coordinates are used to construct
the graph. This common spatial representation forms
the basis of the graph statistics, community,
centrality, spectral, and cell--cell interaction
descriptors.

\noindent\textbf{Cellular composition (5D).}
The relative abundance of each predicted nuclear
phenotype is calculated within each patch. Let $N$
denote the total number of valid nuclei and $N_t$
the number belonging to cell type
$t\in\{1,\ldots,5\}$. The corresponding proportion is
\begin{equation}
	C_t=\frac{N_t}{N},
	\qquad
	\sum_{t=1}^{5}C_t=1.
\end{equation}
These five descriptors quantify the relative
proportions of neoplastic, inflammatory, connective,
dead, and non-neoplastic epithelial nuclei.

\noindent\textbf{Nuclear morphology (30D and 50D).}
Seven morphological descriptors are extracted from
each nuclear contour: area, perimeter, major-axis
length, minor-axis length, eccentricity, solidity,
and ellipse orientation. Major- and minor-axis
lengths and orientation are obtained by fitting
an ellipse to the nuclear contour, while solidity
is defined as the ratio of nuclear area to
convex-hull area.

Global nuclear morphology is summarized using
the mean, standard deviation, skewness, and
kurtosis of the seven descriptors, producing
28 features. The implemented representation
additionally includes the mean and standard
deviation of nuclear area as two repeated
coordinates, resulting in a 30-dimensional
global morphology vector. These coordinates
were retained consistently across all experiments.

For cell-type-specific morphology, the mean and
standard deviation of nuclear area, major-axis
length, minor-axis length, eccentricity, and
solidity are calculated separately for each
of the five predicted cell types. This produces
a 50-dimensional representation. When a cell
type is absent from a patch, its corresponding
morphological descriptors are assigned zero.

\noindent\textbf{Global graph statistics (7D).}
Seven descriptors summarize the spatial graph:
number of vertices, number of edges, graph
density, mean node degree, standard deviation
of node degree, average clustering coefficient,
and transitivity. Together, these measurements
describe cellular abundance, spatial connectivity,
and local neighborhood structure.

\noindent\textbf{Leiden community descriptors (17D).}
The Leiden algorithm \cite{Traag2019Leiden} is
applied to the spatial graph using
modularity-based partitioning to identify
cellular communities.

Eight descriptors characterize general community
structure: number of communities, largest
community size, mean and standard deviation
of community size, modularity, mean spatial
compactness, mean graph diameter, and mean
graph radius. Spatial compactness is calculated
from pairwise Euclidean distances between
nuclei within each community. Community
diameter and radius are computed from
shortest-path distances in the corresponding
connected community subgraphs.

Nine additional descriptors characterize
community composition and spatial relationships:
the mean, standard deviation, and maximum
of community cell-type entropy; proportions
of homotypic and heterotypic communities;
the number and proportion of edges connecting
different communities; and the minimum and
mean centroid distances between communities
containing neoplastic and connective nuclei.

A homotypic community contains nuclei of
only one predicted cell type, whereas a
heterotypic community contains multiple
predicted cell types. Inter-community
proximity is measured using community
centroids rather than individual cell-to-cell
distances.

\noindent\textbf{Neoplastic-community descriptors (8D).}
A separate feature group characterizes
the organization of neoplastic communities.
A pure neoplastic community contains only
neoplastic nuclei, whereas a
neoplastic-containing community includes
at least one neoplastic nucleus and may
also contain other cell types.

The eight descriptors comprise the number
of pure neoplastic communities; mean and
minimum pairwise centroid distances between
pure neoplastic communities; number of
neoplastic-containing communities; mean,
minimum, and standard deviation of pairwise
centroid distances between
neoplastic-containing communities; and
the proportion of all communities that
are pure neoplastic.

Distance-based descriptors are assigned
zero when fewer than two relevant
communities are present. The general
Leiden and neoplastic-community
descriptors are maintained as separate
feature groups to investigate their
predictive contributions independently.

\noindent\textbf{Graph centrality (4D).}
Four centrality measures are computed
for the spatial graph: degree,
betweenness, closeness, and eigenvector
centrality. The mean of each measure
across all vertices forms a
four-dimensional representation
summarizing cellular connectivity
and network organization.

\noindent\textbf{Spectral graph descriptors (3D).}
Three descriptors are obtained from the
eigenvalues of the combinatorial graph
Laplacian
\begin{equation}
	L=D-A,
\end{equation}
where $D$ is the degree matrix and $A$
is the adjacency matrix.

Let
$\lambda_1\leq\lambda_2\leq
\cdots\leq\lambda_N$
denote the Laplacian eigenvalues.
The three spectral descriptors are
the largest eigenvalue $\lambda_N$,
the second-smallest eigenvalue
$\lambda_2$, and the sum of absolute
deviations of the eigenvalues from
their mean:
\begin{equation}
	E_L=
	\sum_{i=1}^{N}
	|\lambda_i-\bar{\lambda}|,
	\qquad
	\bar{\lambda}=
	\frac{1}{N}\sum_{i=1}^{N}\lambda_i.
\end{equation}

The second-smallest eigenvalue
represents algebraic connectivity
and equals zero for a disconnected
graph. The final descriptor
summarizes the dispersion of
the Laplacian spectrum.

\noindent\textbf{Cell--cell interactions (15D).}
Each edge of the spatial graph is
categorized according to the
predicted cell types of its two
incident nuclei. With five
cell types, there are
\begin{equation}
	\binom{5+1}{2}=15
\end{equation}
unique unordered cell-type pairs,
including homotypic and heterotypic
interactions.

The number of graph edges belonging
to each pair defines a 15-dimensional
interaction vector. These descriptors
quantify local spatial relationships
among predicted nuclear populations
within the fixed-distance graph.

The nine feature groups are
concatenated to obtain a
139-dimensional representation
for each eligible image patch.
Non-finite feature values are
replaced with numerical defaults
during feature extraction.

The resulting representation
captures complementary aspects
of tissue architecture, from
individual nuclear characteristics
to spatial connectivity and
multicellular organization.
The feature groups and their
dimensions are summarized in
Table~\ref{tab:features}.

\begin{table}[H]
	\centering
	\caption{Summary of the interpretable
		cellular-to-community features
		extracted from each image patch.}
	\label{tab:features}
	\begin{tabular}{p{3.8cm} p{8.2cm} c}
		\hline
		\textbf{Feature group} &
		\textbf{Descriptors} &
		\textbf{Dimension} \\
		\hline
		
		Cellular composition &
		Proportions of five predicted
		nuclear phenotypes & 5 \\
		\hline
		
		Global nuclear morphology &
		Summary statistics of nuclear
		shape descriptors & 30 \\
		\hline
		
		Per-type nuclear morphology &
		Cell-type-specific morphological
		summary statistics & 50 \\
		\hline
		
		Global graph statistics &
		Nodes, edges, density, degree
		statistics, clustering, and
		transitivity & 7 \\
		\hline
		
		Leiden community descriptors &
		Community size, modularity,
		geometry, phenotype entropy,
		mixing, and inter-community
		proximity & 17 \\
		\hline
		
		Neoplastic-community descriptors &
		Pure and neoplastic-containing
		community counts, centroid
		distances, and proportions & 8 \\
		\hline
		
		Graph centrality &
		Mean degree, betweenness,
		closeness, and eigenvector
		centrality & 4 \\
		\hline
		
		Spectral graph features &
		Laplacian eigenvalues and
		centered Laplacian spectral
		energy & 3 \\
		\hline
		
		Cell--cell interactions &
		Edge counts across 15 unordered
		cell-type pairs & 15 \\
		\hline
		
		\textbf{Total} &
		& \textbf{139} \\
		\hline
	\end{tabular}
\end{table}

\begin{figure}[httb]
		\centering
		\includegraphics[width=1.0\linewidth]{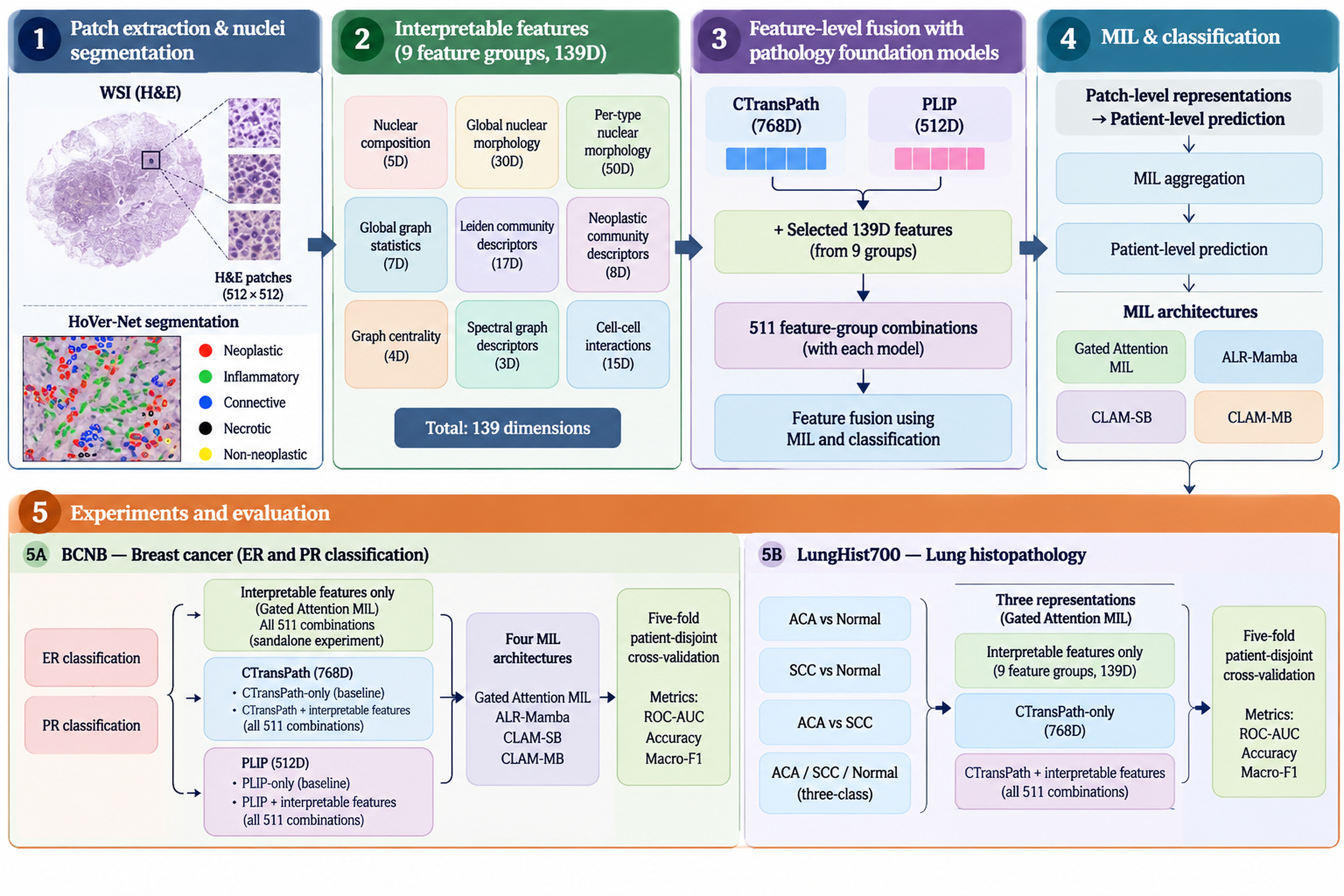}

\caption{Overview of the proposed framework. H\&E image patches are processed using HoVer-Net to obtain nuclear segmentation and cell-type
predictions, from which nine groups of interpretable tissue-architecture descriptors are extracted. The resulting features are evaluated independently and through feature-level fusion with CTransPath (768-dimensional) and PLIP (512-dimensional) embeddings.
For BCNB, all 511 nonempty feature-group combinations are investigated for ER and PR classification, with fusion evaluated across four MIL architectures. For LungHist700, interpretable features, CTransPath embeddings, and their fusion are evaluated using
Gated Attention MIL across four lung histopathology classification tasks. Both datasets are evaluated using five-fold patient-disjoint cross-validation.}

		\label{fig:communityfullframework}
	\end{figure}

\subsection{MIL and Feature-Level Fusion}
\label{subsec:mil_fusion}

Each eligible histopathology patch is represented by a
139-dimensional cellular-to-community feature vector.
For multiple instance learning (MIL), a whole-slide
image (WSI) or histopathology image is represented as
a bag of patch-level instances:
\begin{equation}
	X=\{x_1,x_2,\ldots,x_N\},
	\qquad x_i\in\mathbb{R}^{d},
\end{equation}
where $N$ is the number of patches and $d$ is the
dimension of the selected input representation.
MIL aggregates patch-level information into a
bag-level prediction without requiring individual
patch annotations.

Gated Attention MIL \cite{ilse2018attention} is used
for the standalone interpretable-feature experiments,
including the feature-combination analysis. Its
attention-based aggregation is expressed as
\begin{equation}
	z=\sum_{i=1}^{N}\alpha_i h_i,
	\qquad
	\sum_{i=1}^{N}\alpha_i=1,
\end{equation}
where $h_i$ is the learned representation of patch
$i$, $\alpha_i$ is its normalized attention weight,
and $z$ is the aggregated bag representation used
for classification.

To investigate the complementary predictive value
of interpretable tissue-architecture descriptors,
two pathology foundation models are considered:
CTransPath \cite{wang2022ctranspath} and PLIP
\cite{huang2023plip}. The models independently
extract visual embeddings from the corresponding
H\&E patches, with dimensions of 768 and 512,
respectively.

Let $S\subseteq\{1,\ldots,9\}$ be a nonempty subset
of interpretable feature groups, and let
$c_i^{(S)}$ denote their concatenated descriptors
for patch $i$. For foundation model
$m\in\{\mathrm{CTransPath},\mathrm{PLIP}\}$,
let $v_i^{(m)}\in\mathbb{R}^{d_m}$ denote the
corresponding patch embedding. The fused
representation is defined as
\begin{equation}
	f_i^{(S,m)}
	=
	\left[v_i^{(m)};c_i^{(S)}\right]
	\in\mathbb{R}^{d_m+d_S},
	\qquad
	d_S=\sum_{j\in S}d_j,
\end{equation}
where $d_j$ is the dimension of feature group $j$.
The number of selected groups is
$k=|S|\in\{1,\ldots,9\}$.

The two foundation models are evaluated independently;
their embeddings are not concatenated with each other.
Because the learned embeddings and interpretable
descriptors are obtained from the same H\&E patch,
their concatenation is termed feature-level fusion.

For BCNB, foundation-model-only and fused
representations are evaluated using four MIL
architectures: Gated Attention MIL
\cite{ilse2018attention}, ALR-Mamba
\cite{liu2026alr}, CLAM-SB, and CLAM-MB
\cite{lu2021data}. These architectures provide
different aggregation mechanisms through which
the observed predictive contributions of
interpretable descriptors can be examined.

\subsection{Experimental Design}
\label{subsec:experimental_design}

\noindent\textbf{BCNB experiments.}
Estrogen receptor (ER) and progesterone receptor
(PR) status were evaluated as separate binary
classification tasks using five-fold
patient-disjoint cross-validation. All patches
from a given patient were assigned to the same
fold, preventing patient overlap between the
training and held-out evaluation partitions.
Within each experiment, foundation-model-only
baselines and corresponding fusion configurations
were evaluated using identical patient-level
folds.

The nine interpretable feature groups were
investigated through an exhaustive analysis
of all
\begin{equation}
	\sum_{k=1}^{9}\binom{9}{k}
	=2^9-1=511
\end{equation}
nonempty combinations. Each combination was
first evaluated independently using Gated
Attention MIL without foundation-model embeddings.
For each combination size
$k\in\{1,\ldots,9\}$, the configuration with
the highest mean five-fold ROC-AUC was identified.
The highest-scoring configuration across all
combination sizes was also reported separately
for ER and PR.

Feature-level fusion experiments were then
conducted independently for CTransPath
\cite{wang2022ctranspath} and PLIP
\cite{huang2023plip}. For each foundation
model and MIL architecture, all 511 nonempty
combinations were evaluated by concatenating
the selected interpretable descriptors with
the corresponding patch embeddings. Feature
combinations were not preselected from the
standalone experiments.

For each combination size $k$, the
configuration achieving the highest mean
five-fold ROC-AUC was identified. The
highest-scoring configuration across all
combination sizes was subsequently reported,
together with its corresponding value of $k$.

Consequently, the reported selected-fusion
results represent the maximum observed
cross-validation performance among the
evaluated configurations, rather than
performance representative of every
feature-group combination. Comparisons
with foundation-model-only baselines
therefore describe selected configurations
and do not imply that all combinations
improve classification.

Because feature-group selection and
performance reporting used the same
five-fold cross-validation results,
the reported maxima may be affected
by selection optimism. They should
therefore be interpreted as exploratory
cross-validation findings rather than
independently validated estimates of
the selected configurations'
generalization performance.

Fusion experiments were conducted using
Gated Attention MIL \cite{ilse2018attention},
ALR-Mamba \cite{liu2026alr}, CLAM-SB,
and CLAM-MB \cite{lu2021data} for both
ER and PR classification. ALR-Mamba was
implemented independently following its
published methodological description
\cite{liu2026alr}. Previously published
ALR-Mamba results were included only
for contextual comparison and were
not treated as directly reproduced
experimental results.

\medskip
\noindent\textbf{LungHist700 experiments.}
LungHist700 was used to investigate the
applicability of the tissue-architecture
representation beyond breast cancer.
Images acquired at $20\times$ and
$40\times$ magnification were included
in four classification tasks:
adenocarcinoma (ACA) versus normal,
squamous cell carcinoma (SCC) versus
normal, ACA versus SCC, and three-class
ACA--SCC--normal classification.

The same nine categories of
interpretable descriptors were
extracted for LungHist700.
Three representation types were
investigated: interpretable
descriptors alone, CTransPath
embeddings alone, and their
feature-level fusion. The CTransPath
embeddings comprised 768 dimensions.
Fusion configurations concatenated
the selected interpretable
feature-group descriptors with the
corresponding CTransPath embeddings.

All representations were evaluated
using Gated Attention MIL, with each
histopathology image treated as a
bag of patches. Five-fold
patient-disjoint cross-validation
was performed, ensuring that images
from the same patient, including
different magnifications, remained
within the same fold. An internal
patient-disjoint validation set
was used for model selection and
early stopping.

For each classification task,
the compared representations
were evaluated using identical
patient-level folds. This enabled
comparisons of the observed
predictive performance of
interpretable descriptors,
foundation-model embeddings,
and their selected fusion
configurations.

\medskip
\noindent\textbf{Evaluation metrics.}
ROC-AUC was the primary performance
metric because it evaluates
discrimination across decision
thresholds. Accuracy and macro-F1
were also reported to characterize
threshold-dependent classification
performance. This choice is
consistent with methodological
guidance emphasizing discrimination
metrics and task-appropriate
performance evaluation in medical
prediction and biomedical image
analysis
\cite{lancet_auc_2025,metrics_reloaded_2024}.

\subsection{Implementation Details}
\label{subsec:implementation}

All multiple-instance learning (MIL) models were
implemented using PyTorch. For the primary Gated
Attention MIL experiments, optimization was
performed using Adam with an initial learning
rate of $1\times10^{-3}$ and weight decay of
$1\times10^{-4}$. Binary cross-entropy with
logits was used for BCNB binary classification.

For the Gated Attention MIL experiments, model
training, learning-rate scheduling, early stopping,
and checkpoint selection were performed using
training-side data within each cross-validation
fold. The held-out fold was reserved exclusively
for evaluation and was not used for model
optimization or checkpoint selection.

CTransPath and PLIP embeddings were extracted
using their respective encoder-specific
preprocessing procedures and aligned patch-wise
with the corresponding interpretable descriptors.
The original embedding dimensions were 768 for
CTransPath and 512 for PLIP.

For ALR-Mamba, we used an independent
implementation with a 256-dimensional internal
representation, a region size of 64, and a state
dimension of 16. The model was trained using the
Adam optimizer with a learning rate of
$1\times10^{-4}$, zero weight decay, and
cross-entropy loss. Training was performed for
30 epochs with a batch size of one and a random
seed of 42. The patch-instance sequence was
randomized during training, whereas evaluation
used the stored instance order. The implementation
used five-fold stratified cross-validation, and
models were evaluated after the final training
epoch without early stopping or validation-based
checkpoint selection.

For CLAM-SB and CLAM-MB, we used the publicly
available CLAM implementation
\cite{lu2021data}. These models followed their
respective architecture-specific training
procedures. Within each MIL architecture, the
same model configuration and training protocol
were used for the foundation-model-only baseline
and the corresponding fusion experiments. The
input representation was the principal difference
between these conditions: the baseline used the
original foundation-model embeddings, whereas
fusion used their concatenation with the selected
interpretable feature groups.

For BCNB, classification performance was
assessed using accuracy, precision, recall,
F1-score, and ROC-AUC. A fixed classification
threshold of 0.50 was used for threshold-dependent
binary metrics. LungHist700 performance was
assessed using accuracy, macro-F1, and ROC-AUC.
For the three-class task, ROC-AUC was calculated
using the multiclass one-versus-rest formulation,
and F1-score was macro-averaged across classes.
Results were summarized as mean and standard
deviation across five held-out folds.

Nuclear segmentation and cell phenotyping were
performed using HoVer-Net through TIAToolbox.
Feature extraction was implemented using NumPy,
SciPy, OpenCV, NetworkX, igraph, and Leidenalg.
Model development and evaluation used Python,
PyTorch, and Scikit-learn.

Random seeds were fixed for model training where
applicable. No dedicated Leiden seed was imposed
during the original community-feature extraction.
Consequently, the extracted community descriptors
may exhibit sensitivity to stochastic variation
in community detection.

Experiments were conducted on a workstation
equipped with an Intel Core Ultra 9 285K
processor, 128~GB RAM, and an NVIDIA RTX 5090
GPU with 32~GB memory.

\section{Results and Discussion}

The results are organized into four parts:
(i) predictive performance of the interpretable
cellular-to-community representation,
(ii) its complementary predictive value when
combined with CTransPath and PLIP embeddings,
(iii) associations between tissue-architecture
descriptors and histopathological characteristics,
and (iv) evaluation on LungHist700.
ROC-AUC is used as the primary discrimination
metric. Results are interpreted within their
respective evaluation protocols, without direct
comparisons across experiments employing
different training or evaluation designs.

\subsection{Feature-Combination Ablation}

\begin{figure}[H]
	\centering
	\includegraphics[width=1.0\linewidth]
	{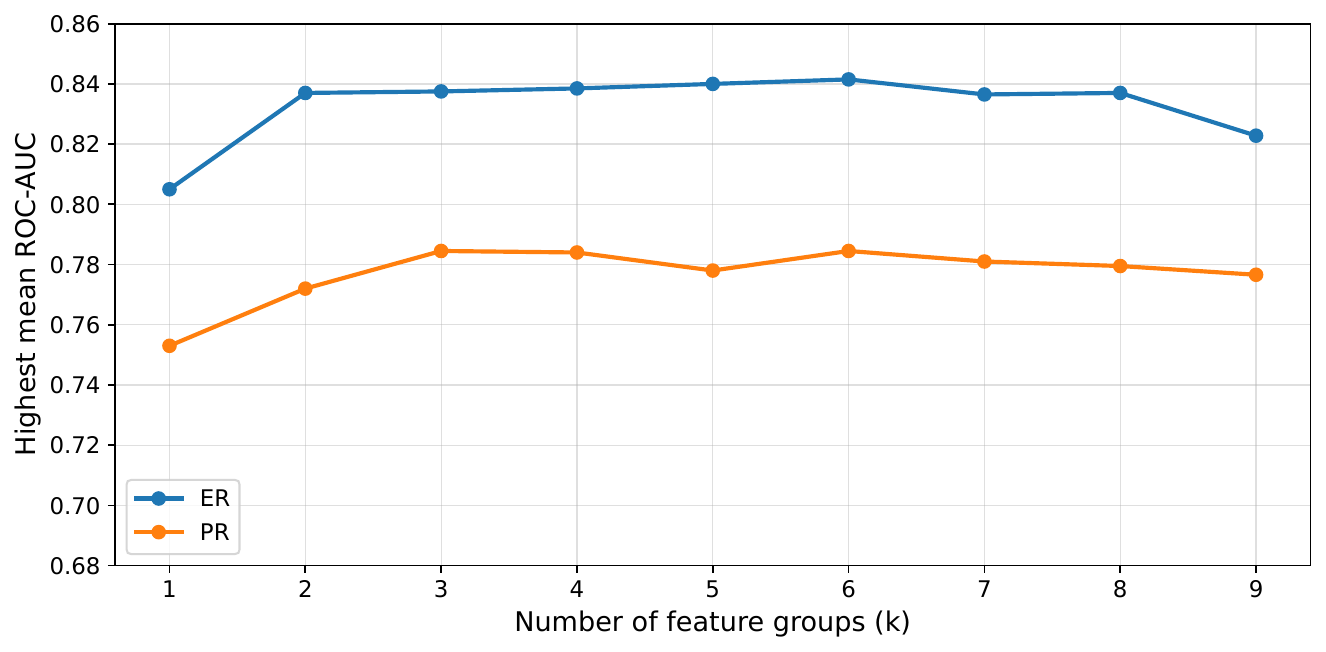}
	
	\caption{Highest observed mean five-fold
		cross-validation ROC-AUC across
		feature-group combination sizes
		($k=1,\ldots,9$) for ER and PR
		classification using Gated Attention
		MIL. Each point represents the
		highest-scoring combination among
		all combinations containing exactly
		$k$ feature groups. The maximum
		observed mean ROC-AUC occurs at
		$k=6$ for ER and $k=3$ for PR,
		rather than at $k=9$.
		These results illustrate the
		dependence of observed predictive
		performance on feature-group
		selection and combination size.}
	
	\label{fig:communityauc1}
\end{figure}

All $2^9-1=511$ nonempty combinations of the nine
interpretable feature groups were evaluated
independently for ER and PR classification using
Gated Attention MIL. Figure~\ref{fig:communityauc1}
shows the highest observed mean five-fold
cross-validation ROC-AUC among combinations
containing exactly $k$ feature groups,
where $k\in\{1,\ldots,9\}$.

The highest observed mean ROC-AUC varied
non-monotonically with the number of selected
feature groups. For ER classification, the
maximum was $0.8415\pm0.0413$ at $k=6$,
whereas PR classification achieved its
maximum of $0.7845\pm0.0365$ at $k=3$.
In comparison, combining all nine feature
groups yielded mean ROC-AUC values of
$0.8228$ for ER and $0.7766$ for PR.
Thus, including all available architectural
descriptors did not produce the highest
observed mean ROC-AUC for either endpoint.

The highest-scoring ER combination comprised
cellular composition, global nuclear morphology,
global graph statistics, Leiden community
descriptors, graph centrality, and spectral
graph features. In contrast, the highest-scoring
PR combination contained cellular composition,
global nuclear morphology, and global graph
statistics. Both configurations included
cellular composition, nuclear morphology,
and global graph properties, while the
highest-scoring ER configuration additionally
incorporated community-level and higher-order
graph descriptors.

These findings suggest that the predictive
contributions of tissue-architecture feature
groups may depend on the classification
endpoint. The inclusion of community,
centrality, and spectral descriptors in the
highest-scoring ER combination is consistent
with a potential contribution from broader
spatial organization, although this result
does not establish the independent importance
or biological mechanism of any individual
feature group.

Importantly, each point in
Figure~\ref{fig:communityauc1} represents
the maximum mean ROC-AUC among multiple
candidate combinations of the same size.
Because the same cross-validation folds
were used for feature-combination selection
and performance reporting, these maxima
may be subject to selection optimism.
Accordingly, the observed configurations
should be interpreted as exploratory
findings rather than independently
validated optimal feature subsets.

	\subsection{Complementarity with Pathology Foundation Models}

We investigated whether interpretable cellular-to-community descriptors
provide additional predictive information when combined with CTransPath
and PLIP embeddings. The two foundation models were evaluated separately
using ALR-Mamba, Gated Attention MIL, CLAM-SB, and CLAM-MB.
All comparisons used the corresponding foundation-model-only baseline
within the same five patient-disjoint folds.

\noindent\textbf{Independent ALR-Mamba reference.}
Because the original ALR-Mamba implementation and complete training
configuration were unavailable, we implemented the framework independently
following its published description \cite{liu2026alr}. As shown in
Table~\ref{tab:alr_reference}, our CTransPath--ALR-Mamba implementation
achieved mean ROC-AUC values of 0.8761 for ER and 0.8035 for PR,
compared with the published values of 0.9078 and 0.8503, respectively.
Differences in implementation, preprocessing, and training may account
for part of this discrepancy. These published values are provided for
context only; our independent implementation is the internally controlled
ALR-Mamba reference for the fusion experiments.

\begin{table}[!htbp]
	\centering
	\caption{Published ALR-Mamba results and results from our independent
		implementation using CTransPath on BCNB \cite{liu2026alr}.
		Values are reported as mean $\pm$ standard deviation.
		The independent implementation provides the controlled
		CTransPath reference for the subsequent fusion experiments.}
	\label{tab:alr_reference}
	
	\small
	\setlength{\tabcolsep}{4pt}
	
	\resizebox{\linewidth}{!}{%
		\begin{tabular}{lcccccc}
			\toprule
			& \multicolumn{3}{c}{\textbf{BCNB-ER}}
			& \multicolumn{3}{c}{\textbf{BCNB-PR}} \\
			\cmidrule(lr){2-4}
			\cmidrule(lr){5-7}
			
			\textbf{Source}
			& \textbf{Accuracy}
			& \textbf{Macro-F1}
			& \textbf{AUC}
			& \textbf{Accuracy}
			& \textbf{Macro-F1}
			& \textbf{AUC} \\
			\midrule
			
			Published \cite{liu2026alr}
			& $\mathbf{86.42 \pm 3.75}$
			& $76.85 \pm 7.01$
			& $\mathbf{90.78 \pm 4.27}$
			& $\mathbf{82.45 \pm 2.55}$
			& $73.80 \pm 5.20$
			& $\mathbf{85.03 \pm 1.00}$ \\
			
			Independent implementation
			& $83.84 \pm 2.32$
			& $\mathbf{80.70 \pm 2.58}$
			& $87.61 \pm 1.40$
			& $79.12 \pm 2.68$
			& $73.86 \pm 2.40$
			& $80.35 \pm 3.13$ \\
			
			\bottomrule
		\end{tabular}%
	}
\end{table}

\noindent\textbf{CTransPath fusion.}
For each MIL architecture and endpoint, all 511 nonempty combinations
of the nine feature groups were evaluated in concatenation with
768-dimensional CTransPath embeddings, without preselection from the
standalone experiments. Table~\ref{tab:fused_features} compares the
highest-scoring fusion configuration with its CTransPath-only baseline.

Selected fusion configurations achieved higher mean ROC-AUC in all eight
comparisons. For ER, the observed differences ranged from 0.13
percentage points (CLAM-MB: 90.90\% to 91.03\%) to 1.84 points
(CLAM-SB: 89.60\% to 91.44\%). The highest ER mean ROC-AUC was
91.44\% $\pm$ 1.43\% under CLAM-SB, using four feature groups.
For PR, the differences ranged from 1.33 points (CLAM-MB:
81.40\% to 82.73\%) to 3.25 points (ALR-Mamba:
80.35\% to 83.60\%). The highest PR mean ROC-AUC was
83.60\% $\pm$ 2.45\% under ALR-Mamba, using three feature groups.

\begin{table}[H]
	\centering
	
	\caption{Feature-level fusion of CTransPath embeddings
		with interpretable tissue-architecture features on BCNB.
		Values are reported as mean $\pm$ standard deviation
		(\%) across five patient-disjoint folds.
		``+ Interpretable'' denotes feature-level concatenation
		with CTransPath embeddings. $k_{\mathrm{ER}}$ and
		$k_{\mathrm{PR}}$ denote the numbers of interpretable
		feature groups in the selected ER and PR fusion
		configurations, respectively. Fusion configurations
		were selected according to the highest mean
		cross-validation ROC-AUC. For each MIL architecture
		and classification task, the higher mean value of
		each evaluation metric between the CTransPath-only
		and fusion configurations is highlighted in bold.}
	
	\label{tab:fused_features}
	
	\setlength{\tabcolsep}{3pt}
	\renewcommand{\arraystretch}{1.15}
	
	\resizebox{\textwidth}{!}{%
		\begin{tabular}{ll cccc cccc}
			\toprule
			
			\textbf{MIL architecture} &
			\textbf{Representation} &
			\multicolumn{4}{c}{\textbf{BCNB-ER}} &
			\multicolumn{4}{c}{\textbf{BCNB-PR}} \\
			
			\cmidrule(lr){3-6}
			\cmidrule(lr){7-10}
			
			& &
			$\boldsymbol{k_{\mathrm{ER}}}$ &
			\textbf{Accuracy} &
			\textbf{Macro-F1} &
			\textbf{AUC} &
			$\boldsymbol{k_{\mathrm{PR}}}$ &
			\textbf{Accuracy} &
			\textbf{Macro-F1} &
			\textbf{AUC} \\
			
			\midrule
			
			\multirow{2}{*}{ALR-Mamba \cite{liu2026alr}}
			& CTransPath
			& --
			& $83.84 \pm 2.32$
			& $\mathbf{80.70 \pm 2.58}$
			& $87.61 \pm 1.40$
			& --
			& $\mathbf{79.12 \pm 2.68}$
			& $\mathbf{73.86 \pm 2.40}$
			& $80.35 \pm 3.13$ \\
			
			& CTransPath + Interpretable
			& 4
			& $\mathbf{85.28 \pm 1.96}$
			& $77.94 \pm 3.97$
			& $\mathbf{89.44 \pm 2.72}$
			& 3
			& $78.35 \pm 1.59$
			& $73.57 \pm 1.48$
			& $\mathbf{83.60 \pm 2.45}$ \\
			
			\midrule
			
			\multirow{2}{*}{Gated Attention MIL
				\cite{ilse2018attention}}
			& CTransPath
			& --
			& $84.55 \pm 2.17$
			& $79.32 \pm 2.62$
			& $90.42 \pm 1.91$
			& --
			& $79.15 \pm 2.73$
			& $73.62 \pm 2.66$
			& $81.52 \pm 2.83$ \\
			
			& CTransPath + Interpretable
			& 5
			& $\mathbf{86.16 \pm 1.25}$
			& $\mathbf{81.03 \pm 1.54}$
			& $\mathbf{90.80 \pm 1.68}$
			& 3
			& $\mathbf{80.85 \pm 4.21}$
			& $\mathbf{74.43 \pm 4.74}$
			& $\mathbf{83.03 \pm 3.00}$ \\
			
			\midrule
			
			\multirow{2}{*}{CLAM-SB \cite{lu2021data}}
			& CTransPath
			& --
			& $84.83 \pm 1.24$
			& $75.67 \pm 3.26$
			& $89.60 \pm 2.37$
			& --
			& $\mathbf{79.81 \pm 1.49}$
			& $\mathbf{69.77 \pm 2.38}$
			& $81.14 \pm 3.44$ \\
			
			& CTransPath + Interpretable
			& 4
			& $\mathbf{86.26 \pm 2.45}$
			& $\mathbf{77.69 \pm 6.63}$
			& $\mathbf{91.44 \pm 1.43}$
			& 3
			& $78.96 \pm 1.60$
			& $67.79 \pm 5.17$
			& $\mathbf{82.82 \pm 2.39}$ \\
			
			\midrule
			
			\multirow{2}{*}{CLAM-MB \cite{lu2021data}}
			& CTransPath
			& --
			& $85.97 \pm 1.71$
			& $78.42 \pm 2.73$
			& $90.90 \pm 1.70$
			& --
			& $79.72 \pm 2.60$
			& $70.97 \pm 3.33$
			& $81.40 \pm 3.03$ \\
			
			& CTransPath + Interpretable
			& 3
			& $\mathbf{87.39 \pm 2.07}$
			& $\mathbf{81.24 \pm 3.43}$
			& $\mathbf{91.03 \pm 1.68}$
			& 2
			& $\mathbf{80.47 \pm 2.59}$
			& $\mathbf{71.52 \pm 4.37}$
			& $\mathbf{82.73 \pm 2.61}$ \\
			
			\bottomrule
		\end{tabular}%
	}
\end{table}

\noindent\textbf{PLIP fusion.}
The corresponding experiments were conducted independently with
512-dimensional PLIP embeddings. As shown in
Table~\ref{tab:plip_fused_features}, the selected fusion configurations
achieved higher mean ROC-AUC than their PLIP-only baselines in all eight
ER and PR comparisons.

For ER, observed differences ranged from 1.10 percentage points
(CLAM-MB: 86.45\% to 87.55\%) to 2.45 points
(CLAM-SB: 85.46\% to 87.91\%). The highest mean ER ROC-AUC was
87.91\% $\pm$ 0.96\% with CLAM-SB and four feature groups.
For PR, differences ranged from 0.76 points (Gated Attention MIL:
79.03\% to 79.79\%) to 2.40 points (ALR-Mamba:
74.62\% to 77.02\%). The highest mean PR ROC-AUC was
80.15\% $\pm$ 1.34\% with CLAM-MB and three feature groups.

Across both foundation models, changes in accuracy and macro-F1 were
less consistent than changes in ROC-AUC. For example, CTransPath fusion
with ALR-Mamba improved ER accuracy but reduced macro-F1, while PLIP
fusion with Gated Attention MIL improved ER ROC-AUC despite lower
accuracy and macro-F1. Thus, higher discrimination did not necessarily
translate into better threshold-dependent classification performance.

\begin{table}[htbp]
	\centering
	
	\caption{Feature-level fusion of PLIP embeddings
		with interpretable tissue-architecture features
		on BCNB. Values are reported as mean $\pm$
		standard deviation (\%) across five patient-disjoint
		folds. ``+ Interpretable'' denotes feature-level
		concatenation with PLIP embeddings.
		$k_{\mathrm{ER}}$ and $k_{\mathrm{PR}}$ indicate
		the numbers of interpretable feature groups
		in the selected ER and PR fusion configurations,
		respectively. Fusion configurations were selected
		according to the highest mean cross-validation
		ROC-AUC. For each MIL architecture and
		classification task, the higher mean value
		of each evaluation metric between the PLIP-only
		and fusion configurations is highlighted in bold.}
	
	\label{tab:plip_fused_features}
	
	\setlength{\tabcolsep}{3pt}
	\renewcommand{\arraystretch}{1.15}
	
	\resizebox{\textwidth}{!}{%
		\begin{tabular}{ll cccc cccc}
			\toprule
			
			\textbf{MIL architecture} &
			\textbf{Representation} &
			\multicolumn{4}{c}{\textbf{BCNB-ER}} &
			\multicolumn{4}{c}{\textbf{BCNB-PR}} \\
			
			\cmidrule(lr){3-6}
			\cmidrule(lr){7-10}
			
			& &
			$\boldsymbol{k_{\mathrm{ER}}}$ &
			\textbf{Accuracy} &
			\textbf{Macro-F1} &
			\textbf{AUC} &
			$\boldsymbol{k_{\mathrm{PR}}}$ &
			\textbf{Accuracy} &
			\textbf{Macro-F1} &
			\textbf{AUC} \\
			
			\midrule
			
			\multirow{2}{*}{ALR-Mamba \cite{liu2026alr}}
			& PLIP
			& --
			& $\mathbf{82.86 \pm 0.51}$
			& $72.40 \pm 2.93$
			& $82.47 \pm 2.18$
			& --
			& $76.32 \pm 7.37$
			& $69.20 \pm 5.25$
			& $74.62 \pm 3.77$ \\
			
			& PLIP + Interpretable
			& 3
			& $82.78 \pm 0.81$
			& $\mathbf{74.18 \pm 2.82}$
			& $\mathbf{84.41 \pm 2.73}$
			& 3
			& $\mathbf{78.31 \pm 1.39}$
			& $\mathbf{71.84 \pm 3.08}$
			& $\mathbf{77.02 \pm 6.15}$ \\
			
			\midrule
			
			\multirow{2}{*}{Gated Attention MIL
				\cite{ilse2018attention}}
			& PLIP
			& --
			& $\mathbf{80.66 \pm 2.13}$
			& $\mathbf{74.16 \pm 1.30}$
			& $85.44 \pm 1.10$
			& --
			& $77.35 \pm 2.41$
			& $71.24 \pm 2.96$
			& $79.03 \pm 1.42$ \\
			
			& PLIP + Interpretable
			& 4
			& $79.81 \pm 2.33$
			& $74.04 \pm 2.77$
			& $\mathbf{86.67 \pm 2.16}$
			& 3
			& $\mathbf{78.96 \pm 3.64}$
			& $\mathbf{71.98 \pm 3.67}$
			& $\mathbf{79.79 \pm 2.09}$ \\
			
			\midrule
			
			\multirow{2}{*}{CLAM-SB \cite{lu2021data}}
			& PLIP
			& --
			& $83.32 \pm 0.92$
			& $72.88 \pm 3.41$
			& $85.46 \pm 2.37$
			& --
			& $80.38 \pm 2.41$
			& $70.06 \pm 3.48$
			& $78.20 \pm 1.38$ \\
			
			& PLIP + Interpretable
			& 4
			& $\mathbf{84.17 \pm 1.15}$
			& $\mathbf{75.66 \pm 0.98}$
			& $\mathbf{87.91 \pm 0.96}$
			& 2
			& $\mathbf{80.76 \pm 2.19}$
			& $\mathbf{71.97 \pm 3.04}$
			& $\mathbf{79.72 \pm 1.86}$ \\
			
			\midrule
			
			\multirow{2}{*}{CLAM-MB \cite{lu2021data}}
			& PLIP
			& --
			& $83.41 \pm 1.79$
			& $74.52 \pm 2.33$
			& $86.45 \pm 1.30$
			& --
			& $\mathbf{81.04 \pm 2.26}$
			& $\mathbf{72.19 \pm 4.63}$
			& $79.01 \pm 1.52$ \\
			
			& PLIP + Interpretable
			& 3
			& $\mathbf{84.36 \pm 2.16}$
			& $\mathbf{75.46 \pm 2.74}$
			& $\mathbf{87.55 \pm 1.11}$
			& 3
			& $80.85 \pm 2.72$
			& $69.55 \pm 6.29$
			& $\mathbf{80.15 \pm 1.34}$ \\
			
			\bottomrule
		\end{tabular}%
	}
\end{table}

\subsection{Cellular-to-Community Interpretability}

\begin{figure}[httb]
	\centering
	\includegraphics[width=1.0\linewidth, height=0.5\textheight, keepaspectratio]
	{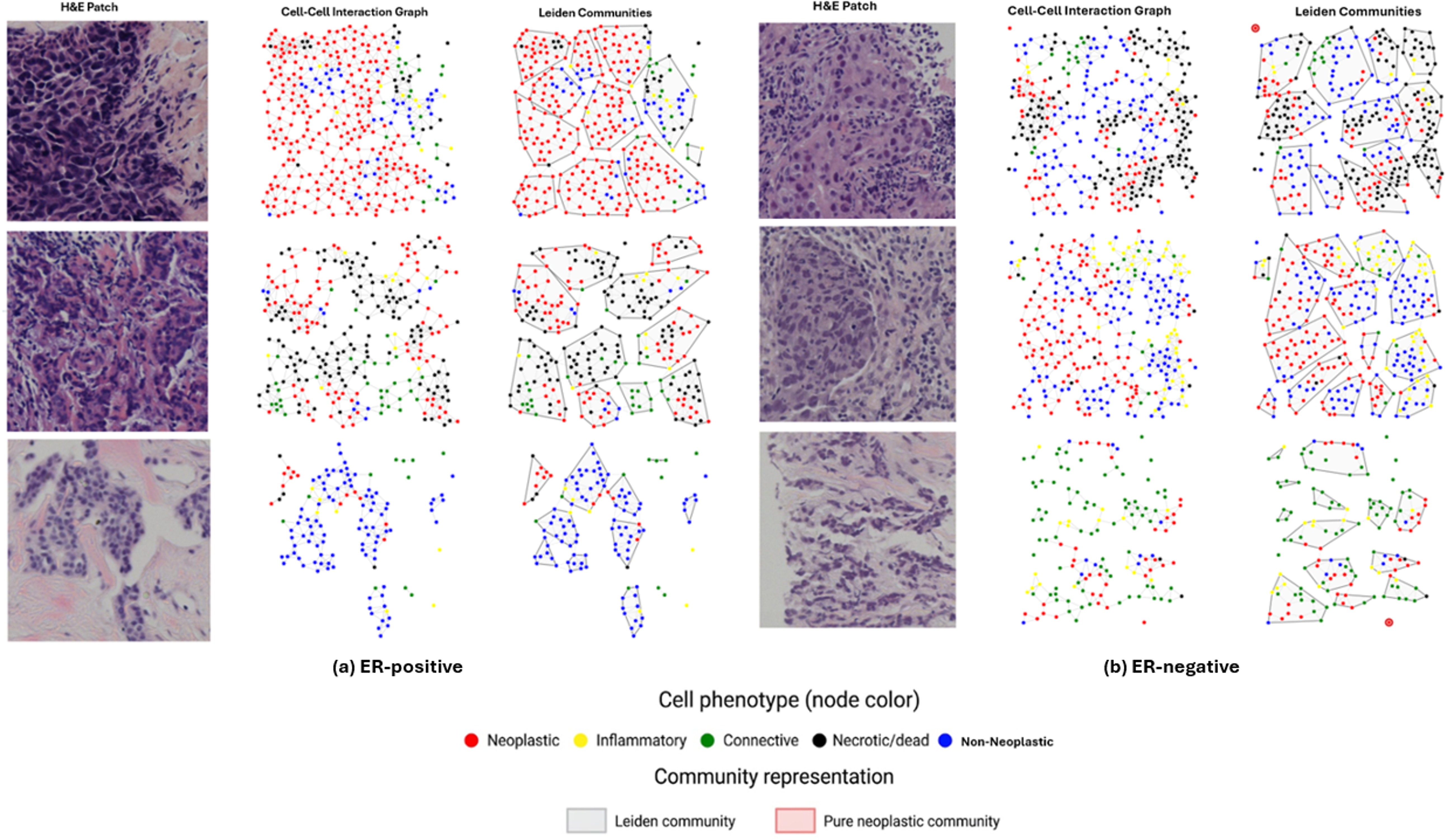}
	
	\caption{Qualitative visualization of the
		cellular-to-community representation in
		objectively selected ER-positive and
		ER-negative cases. Patients were ranked
		by their distance from the corresponding
		ER-group median of the patient-level
		mean number of pure neoplastic communities.
		The three nearest patients per group
		were selected, and the displayed patch
		for each patient was chosen according
		to its proximity to that patient's
		median patch-level value.
		For each patch, the H\&E image,
		cell--cell interaction graph, and
		corresponding Leiden community
		partition are shown. Node colors
		indicate predicted nuclear phenotypes:
		red, neoplastic; yellow, inflammatory;
		green, connective; black, dead;
		and blue, non-neoplastic epithelial.
		Gray outlines indicate Leiden
		communities, while thicker red
		outlines identify pure neoplastic
		communities containing exclusively
		neoplastic nuclei. The examples
		illustrate the extracted spatial
		representation and do not constitute
		population-level evidence of
		ER-associated architectural differences.}
	
	\label{fig:graph_community_qualitative}
\end{figure}

Figure~\ref{fig:graph_community_qualitative}
illustrates the transformation of nuclear phenotypes
detected in H\&E-stained tissue into spatial graphs
and higher-order cellular communities. Previous
breast cancer studies have demonstrated the
relevance of cellular morphology, intercellular
relationships, and multicellular spatial organization
to tissue characterization
\cite{jackson2020singlecell,danenberg2022tme,zhao2023scmtop}.

To avoid subjective visual selection, ER-positive
and ER-negative cases were selected using a
predefined median-based procedure based on the
patient-level mean number of pure neoplastic
communities. Three patients nearest to the
corresponding ER-group median were retained,
and one patch nearest to each selected patient's
median patch-level value was displayed.

The visualizations distinguish cellular composition
from spatial tissue organization. While cellular
composition describes the relative abundance of
different nuclear phenotypes, the spatial graph
additionally represents their local connectivity.
Leiden community detection further identifies
groups of spatially connected nuclei, including
phenotypically heterogeneous communities and
pure neoplastic communities containing exclusively
neoplastic nuclei.

This distinction is relevant to breast cancer
histopathology because spatial relationships
among tumor, stromal, and immune populations
may contain information not captured by their
relative abundances alone. Previous studies
have associated spatial patterns involving
tumor cells, lymphocytes, and fibroblasts
with molecular and clinical characteristics
in ER-positive breast cancer
\cite{nederlof2022spatial}. Spatial
heterogeneity in tumor and immune cell
distributions has also been investigated
in ER-negative disease
\cite{nawaz2015spatial}.

The examples in Figure~\ref{fig:graph_community_qualitative}
demonstrate how the proposed representation
captures local cellular connectivity and
multicellular organization. However, differences
visible between individual patches cannot
establish systematic ER-associated spatial
patterns. The figure is therefore presented as a qualitative
illustration of the representation, while
patient-level associations between tissue-architecture
descriptors and ER status are examined quantitatively
below.

To investigate whether the interpretable
representation captures measurable tissue
characteristics associated with ER status,
we conducted a patient-level analysis
across the complete 139-dimensional
feature space. Patch-level descriptors
were mean-pooled within each patient,
and ER-negative and ER-positive groups
were compared using two-sided
Mann--Whitney $U$ tests.

Features with fewer than five
non-missing patients in either group
or no variation across the cohort
were excluded. Multiple comparisons
were controlled using the
Benjamini--Hochberg false-discovery-rate
(FDR) procedure at a threshold of
$0.05$. Rank-biserial correlation
($r_{\mathrm{rb}}$) quantified the
direction and magnitude of the
observed differences, with positive
values indicating larger values
in ER-negative patients and negative
values indicating larger values
in ER-positive patients.

\begin{figure}[!t]
	\centering
	\includegraphics[
	width=0.95\linewidth,
	height=0.70\textheight,
	keepaspectratio
	]{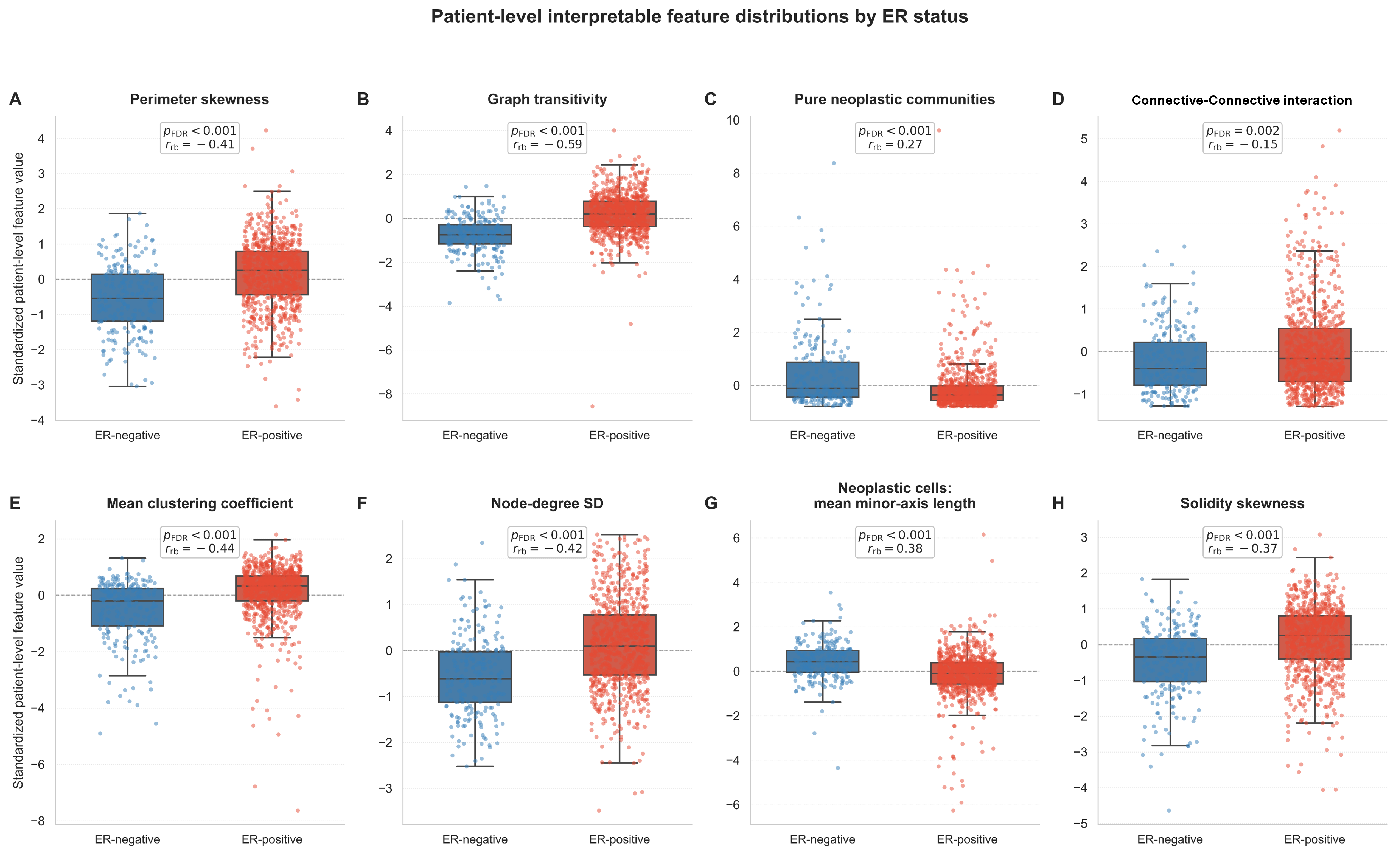}
	
	\caption{Patient-level distributions
		of eight representative
		FDR-supported tissue-architecture
		descriptors associated with
		ER status: (A) perimeter skewness,
		(B) graph transitivity,
		(C) pure neoplastic communities,
		(D) connective--connective
		interaction counts,
		(E) mean clustering coefficient,
		(F) node-degree standard
		deviation, (G) neoplastic-cell
		mean minor-axis length, and
		(H) solidity skewness.
		Patch-level features were
		mean-pooled within each patient,
		and statistical testing used
		the original patient-level
		values. Z-standardization
		was applied only for
		visualization. ER-negative
		($n=227$) and ER-positive
		($n=831$) patients were
		compared using two-sided
		Mann--Whitney $U$ tests
		with Benjamini--Hochberg
		correction across all
		tested features.
		Adjusted $p$-values and
		rank-biserial effect sizes
		are displayed in each
		panel. Positive
		$r_{\mathrm{rb}}$ indicates
		larger values in ER-negative
		patients, whereas negative
		values indicate larger
		values in ER-positive
		patients.}
	
	\label{fig:global_morph_bar}
\end{figure}

Figure~\ref{fig:global_morph_bar}
presents eight representative
FDR-supported features spanning
nuclear morphology, graph
connectivity, cellular communities,
and cell--cell interactions.
Features were selected to
illustrate associations across
the cellular-to-community hierarchy
rather than solely on the basis
of statistical significance.
All statistical tests were
performed on the original
patient-level feature values;
z-standardization was applied
only for visualization.

\noindent\textbf{Nuclear morphology.}
ER-negative patients exhibited
larger neoplastic-cell mean
minor-axis length (Fig.~\ref{fig:global_morph_bar}G;
$r_{\mathrm{rb}}=0.38$).
Previous studies have reported
relationships between nuclear
morphology and ER status,
supporting the relevance of
nuclear shape and size to
breast cancer characterization
\cite{rawat2018ermorph,naik2020receptornet}.

In contrast, perimeter skewness
(panel A; $r_{\mathrm{rb}}=-0.41$)
and solidity skewness
(panel H; $r_{\mathrm{rb}}=-0.37$)
were higher in ER-positive patients.
These descriptors characterize
the asymmetry of nuclear
morphological distributions
rather than differences in
their average values.
Although nuclear perimeter
and solidity are established
quantitative morphological
measurements
\cite{rawat2018ermorph,zhao2023scmtop},
the specific ER-associated
directions of their skewness
observed here are interpreted
as cohort-level findings.

\noindent\textbf{Spatial graph organization.}
ER-positive patients exhibited
higher graph transitivity
(panel B; $r_{\mathrm{rb}}=-0.59$),
mean clustering coefficient
(panel E; $r_{\mathrm{rb}}=-0.44$),
and node-degree standard deviation
(panel F; $r_{\mathrm{rb}}=-0.42$).

Transitivity and clustering
coefficient quantify the
prevalence of locally connected
cellular neighborhoods, whereas
node-degree variability describes
heterogeneity in the number of
spatial connections per nucleus.
Together, these associations
suggest differences in local
cellular connectivity between
ER groups within the
fixed-distance graph
representation.

Previous spatial studies have
demonstrated the relevance
of multicellular organization
to breast cancer tissue
phenotypes
\cite{jackson2020singlecell,danenberg2022tme,zhao2023scmtop}.
However, these studies do
not independently establish
the specific directions
observed for the graph
descriptors in the present
cohort.

\noindent\textbf{Cellular communities.}
ER-negative patients showed
a higher mean number of pure
neoplastic communities per
patch (panel C;
$r_{\mathrm{rb}}=0.27$).
This descriptor counts
communities composed
exclusively of nuclei
predicted as neoplastic,
thereby characterizing
one aspect of the spatial
organization of neoplastic
cell populations.

Unlike global connectivity
statistics, community-level
descriptors summarize
groups of spatially related
cells. Previous investigations
have demonstrated the
relevance of multicellular
communities and spatial
tissue organization to
breast cancer phenotypes
\cite{jackson2020singlecell,danenberg2022tme,javed2020cellular}.
The observed ER-associated
direction remains specific
to the present cohort and
the adopted graph and
community construction.

\noindent\textbf{Cell--cell interactions.}
Connective--connective
interaction counts were
higher in ER-positive
patients (panel D;
$r_{\mathrm{rb}}=-0.15$).
This was the smallest
effect among the eight
displayed descriptors.

Previous studies have
highlighted the relevance
of stromal organization
and fibroblast-associated
spatial patterns in
breast cancer
\cite{nederlof2022spatial,danenberg2022tme}.
However, the HoVer-Net
connective category does
not provide a
fibroblast-specific
annotation. Furthermore,
the descriptor counts
spatial graph edges
between nuclei classified
as connective, rather
than directly measuring
biological cell--cell
interactions. Accordingly,
this finding is interpreted
as a relatively small
association with local
connective-cell spatial
organization, without
inferring a specific
stromal mechanism.

Overall, Figure~\ref{fig:global_morph_bar}
demonstrates that measurable
ER-associated differences
occur across several
levels of the proposed
representation, including
nuclear morphology,
local graph connectivity,
neoplastic-community
organization, and
connective-cell spatial
relationships.

These findings support
the interpretability of
the cellular-to-community
representation by linking
individual descriptors
to defined histological
properties. However,
the observed associations
are observational and
do not establish causal
mechanisms, independent
predictive contributions,
or generalizability
beyond the evaluated
cohort.

	\FloatBarrier

\subsection{Evaluation on LungHist700}
\label{subsec:lunghist_results}

The cellular-to-community representation was
further evaluated on LungHist700 to investigate
its applicability beyond breast cancer.
All experiments employed Gated Attention MIL
with five-fold patient-disjoint cross-validation.

When evaluated independently of foundation-model
embeddings, the interpretable descriptors
achieved mean ROC-AUC values of
$0.9896\pm0.0089$ for SCC versus normal
and $0.9809\pm0.0160$ for ACA versus normal.
Discrimination was lower for ACA versus SCC
($0.7158\pm0.1103$), while three-class
ACA--SCC--normal classification achieved
$0.8470\pm0.0306$. Thus, the standalone
representation exhibited higher observed
discrimination for malignant-versus-normal
comparisons than for distinguishing the two
malignant histological subtypes.

To investigate potential predictive
complementarity, interpretable descriptors
were subsequently combined with
768-dimensional CTransPath embeddings
through feature-level concatenation.
Feature-group combinations were evaluated
independently of the standalone experiments,
without preliminary selection based on
standalone classification performance.
For each task, the configuration achieving
the highest mean five-fold ROC-AUC was
selected for reporting.

Table~\ref{tab:lunghist700_fusion}
compares the CTransPath-only baseline
with the corresponding selected fusion
configuration for each classification task.

\begin{table}[H]
	\centering
	\caption{Classification performance of
		CTransPath embeddings alone and after
		feature-level fusion with interpretable
		cellular-to-community descriptors on
		LungHist700 using Gated Attention MIL.
		Values are mean $\pm$ standard deviation
		across five patient-disjoint folds.
		Fusion configurations were selected
		according to the highest mean
		cross-validation ROC-AUC. Bold values
		indicate the higher mean performance
		within each classification task.}
	\label{tab:lunghist700_fusion}
	
	\small
	\setlength{\tabcolsep}{3.5pt}
	
	\begin{tabularx}{\linewidth}{l X c c c}
		\toprule
		
		\textbf{Classification task} &
		\textbf{Representation} &
		\textbf{Accuracy} &
		\textbf{Macro-F1} &
		\textbf{ROC-AUC} \\
		
		\midrule
		
		ACA vs Normal &
		CTransPath &
		$0.7547 \pm 0.0782$ &
		$0.7027 \pm 0.0829$ &
		$0.8564 \pm 0.0856$ \\
		
		&
		CTransPath + interpretable ($k=4$) &
		$\mathbf{0.8505 \pm 0.0582}$ &
		$\mathbf{0.8284 \pm 0.0517}$ &
		$\mathbf{0.9231 \pm 0.0511}$ \\
		
		\addlinespace
		
		SCC vs Normal &
		CTransPath &
		$0.8539 \pm 0.0538$ &
		$0.8301 \pm 0.0657$ &
		$0.9426 \pm 0.0412$ \\
		
		&
		CTransPath + interpretable ($k=7$) &
		$\mathbf{0.8911 \pm 0.0770}$ &
		$\mathbf{0.8753 \pm 0.0924}$ &
		$\mathbf{0.9681 \pm 0.0309}$ \\
		
		\addlinespace
		
		ACA vs SCC &
		CTransPath &
		$0.6336 \pm 0.1248$ &
		$0.6222 \pm 0.1311$ &
		$0.7137 \pm 0.1165$ \\
		
		&
		CTransPath + interpretable ($k=8$) &
		$\mathbf{0.7042 \pm 0.0749}$ &
		$\mathbf{0.6969 \pm 0.0706}$ &
		$\mathbf{0.7967 \pm 0.0750}$ \\
		
		\addlinespace
		
		ACA vs SCC vs Normal &
		CTransPath &
		$0.6214 \pm 0.0580$ &
		$0.6146 \pm 0.0434$ &
		$0.8132 \pm 0.0277$ \\
		
		&
		CTransPath + interpretable ($k=6$) &
		$\mathbf{0.6717 \pm 0.0758}$ &
		$\mathbf{0.6701 \pm 0.0792}$ &
		$\mathbf{0.8435 \pm 0.0516}$ \\
		
		\bottomrule
	\end{tabularx}
	
	\vspace{0.4em}
	
	\begin{minipage}{\linewidth}
		\footnotesize
		ACA: adenocarcinoma;
		SCC: squamous cell carcinoma.
		The value of $k$ denotes the number
		of interpretable feature groups in
		the selected fusion configuration.
		Feature-group selection was performed
		independently of the standalone
		experiments. Baseline and fusion
		configurations were evaluated using
		identical patient-disjoint folds.
		Three-class ROC-AUC was computed
		using the multiclass one-versus-rest
		formulation.
	\end{minipage}
\end{table}

The selected fusion configurations achieved
higher mean ROC-AUC, accuracy, and macro-F1
than the corresponding CTransPath-only
baselines across all four tasks.
Mean ROC-AUC differences were $+0.0667$
for ACA versus normal, $+0.0255$ for
SCC versus normal, $+0.0830$ for ACA
versus SCC, and $+0.0303$ for
three-class classification.

The highest-scoring configurations contained
four, seven, eight, and six interpretable
feature groups, respectively. This variation
indicates that the selected architectural
feature combinations differed across
classification tasks, although the
reported group counts alone do not
establish the importance of individual
feature groups.

The largest observed ROC-AUC difference
occurred for ACA-versus-SCC classification,
where the selected fusion configuration
achieved $0.7967\pm0.0750$, compared with
$0.7137\pm0.1165$ for CTransPath alone.
Accuracy increased from $0.6336$ to
$0.7042$, while macro-F1 increased from
$0.6222$ to $0.6969$.

This comparison is particularly relevant
because ACA and SCC are both malignant
lung histological subtypes. Distinguishing
them requires discrimination between
different tumor morphologies and patterns
of tissue organization rather than
simply separating malignant from normal
tissue. The observed results are
consistent with the possibility that
explicit cellular and spatial
architectural descriptors capture
predictive information not fully
represented by CTransPath embeddings
in the evaluated setting.

Nevertheless, the standalone
interpretable representation achieved
a mean ROC-AUC of
$0.7158\pm0.1103$ for ACA versus SCC,
and the observed fusion result does
not establish the independent
contribution of any particular
architectural descriptor. Furthermore,
the selected configuration represents
the highest-scoring combination
identified during the feature search,
rather than a configuration evaluated
on an independent test cohort.

Overall, the LungHist700 experiments
demonstrate the applicability of the
cellular-to-community representation
to pulmonary histopathological
classification within the evaluated
dataset. The standalone descriptors
showed substantial discrimination
for malignant-versus-normal tasks,
while selected fusion configurations
achieved higher mean classification
performance than CTransPath alone
across all four tasks.

These findings suggest potential
complementary predictive value of
explicit tissue-architecture
descriptors beyond breast cancer.
However, the relatively small cohort
of 45 patients, the variability
across cross-validation folds,
and feature-group selection based
on the same folds used for
performance reporting limit
the strength of this conclusion.
The observed differences should
therefore be interpreted as
exploratory findings requiring
independent validation.

\subsection{Discussion}

The findings of this study suggest that explicit
measurements of cellular and multicellular tissue
architecture may provide predictive information
complementary to pathology foundation-model
representations. Three principal observations
support this interpretation: the dependence of
standalone performance on feature-group
composition, the higher mean ROC-AUC achieved
by selected fusion configurations across
different foundation models and MIL architectures,
and the identification of measurable
ER-associated architectural characteristics.

The feature-combination analysis demonstrated
that the highest observed discrimination did
not require all nine feature groups. Cellular
composition, global nuclear morphology, and
global graph statistics appeared in the
highest-scoring standalone configurations
for both ER and PR, whereas community,
centrality, and spectral descriptors were
included selectively. This suggests that
the predictive relevance of architectural
measurements depends on the classification
endpoint and the combination in which they
are evaluated. However, inclusion in a
highest-scoring combination does not establish
the independent contribution of an individual
feature group.

Feature-level fusion provided further
evidence of potential predictive
complementarity. Across both CTransPath
and PLIP and all four MIL architectures,
the selected fusion configurations achieved
higher mean ROC-AUC than their corresponding
foundation-model-only baselines for ER
and PR classification. The highest
CTransPath fusion ROC-AUC values were
0.9144 for ER using CLAM-SB and 0.8360
for PR using ALR-Mamba. The corresponding
highest PLIP fusion values were 0.8791
for ER using CLAM-SB and 0.8015 for PR
using CLAM-MB.

These results indicate that the observed
contribution of interpretable descriptors
was not restricted to a single foundation
model or aggregation architecture.
Nevertheless, the magnitude of the
ROC-AUC differences varied across
experiments, and accuracy and macro-F1
did not improve consistently. Thus,
higher discrimination across decision
thresholds did not necessarily translate
into better classification performance
at the selected operating threshold.

An important distinction is that the
standalone predictive performance of
a representation does not necessarily
determine its additional value in fusion.
This was particularly evident in the
LungHist700 experiments. Although the
interpretable representation achieved
substantially higher standalone
discrimination for malignant-versus-normal
tasks than for ACA versus SCC, the
ACA-versus-SCC comparison exhibited
the largest observed ROC-AUC difference
between selected fusion and CTransPath
alone (0.7967 versus 0.7137).

This finding is relevant because
distinguishing ACA from SCC requires
discrimination between two malignant
histological subtypes rather than
between malignant and normal tissue.
The result is consistent with the
possibility that explicit cellular
and spatial architectural descriptors
capture information not fully represented
by the learned visual embeddings in
the evaluated setting. However,
the experiment does not establish
which individual descriptors account
for the observed difference.

The interpretability analyses provided
a complementary perspective on the
representation. Patient-level
associations with ER status were
identified across nuclear morphology,
graph connectivity, neoplastic-community
organization, and connective-cell
spatial relationships. The qualitative
visualizations further illustrated
how these measurements correspond
to cellular neighborhoods and
multicellular communities in
histopathological tissue.

Together, these analyses demonstrate
that the representation contains
explicitly defined descriptors
whose values can be related to
measurable histological properties.
However, statistical associations
with ER status do not establish
biological causation, and the
association analysis does not
demonstrate that the same descriptors
were responsible for the observed
fusion performance differences.

Several methodological limitations
should be considered. First,
the extracted features depend on
nuclear segmentation and cell-type
classification accuracy. Errors
in nuclear boundaries, centroid
localization, or predicted phenotypes
may propagate into morphological,
graph-based, community, and
interaction descriptors.

Second, the spatial graph was
constructed using a fixed
40-pixel distance threshold,
and community detection relied
on Leiden partitioning.
Consequently, the extracted
architectural measurements may
be sensitive to graph-construction
parameters, image resolution,
and community-detection variability.
These factors were not
systematically evaluated in
the present study.

Third, the exhaustive search
over 511 feature-group combinations
introduces a risk of selection
optimism. Because the same
cross-validation results were
used to select and report
the highest-scoring configurations,
their reported ROC-AUC values
may overestimate performance
on independent data. Moreover,
formal paired statistical testing
was not performed, and the
observed differences should
not be interpreted as
statistically established
improvements.

Finally, although LungHist700
provided an additional
histopathological evaluation
setting, its cohort comprised
only 45 patients. The results
therefore require confirmation
in larger, independent cohorts
before conclusions about
generalizability across
institutions, staining
conditions, or patient
populations can be drawn.

Future investigations should
prioritize independent validation,
nested or otherwise
selection-independent evaluation
of feature combinations, and
robustness analyses addressing
segmentation uncertainty,
spatial graph parameters,
community detection, and
staining variation.

Overall, the results suggest
that interpretable
cellular-to-community
descriptors may complement
learned histopathological
representations while
providing measurable
characterizations of
tissue architecture.
Their observed predictive
contributions depend on
the classification endpoint,
foundation model, and
MIL architecture, and
require independent
validation.

\section{Conclusion}

This study investigated the predictive
value and potential complementarity
of interpretable tissue-architecture
descriptors alongside pathology
foundation-model representations.
A 139-dimensional
cellular-to-community
representation was constructed
from nine feature groups
characterizing nuclear
composition and morphology,
spatial connectivity,
community organization,
graph centrality,
spectral properties,
and cell--cell interactions.

On BCNB, exhaustive evaluation
of 511 feature-group combinations
showed that the highest observed
ER and PR discrimination depended
on feature-group composition
rather than simply the number
of descriptors included.
Selected feature-level fusion
configurations achieved higher
mean ROC-AUC than CTransPath-
and PLIP-only baselines
across four MIL architectures
for both biomarkers, although
accuracy and macro-F1 did
not improve consistently.

Patient-level analyses
identified ER-associated
differences in nuclear
morphology, graph connectivity,
neoplastic-community organization,
and connective-cell spatial
relationships, demonstrating
the ability of the representation
to describe measurable
histopathological characteristics.

Evaluation on LungHist700
extended the investigation
to pulmonary histopathology.
Selected CTransPath fusion
configurations achieved higher
mean ROC-AUC across four
classification tasks, with
the largest observed difference
for ACA versus SCC
(0.7967 versus 0.7137
for CTransPath alone).

These findings suggest that
explicit cellular and
multicellular architectural
descriptors may provide
predictive information
complementary to learned
visual representations.
However, the selected
configurations require
independent validation,
particularly because
feature-group selection
and performance reporting
used the same cross-validation
folds. The observed differences
were not subjected to formal
paired significance testing,
and robustness to segmentation
errors, graph parameters,
and staining variation
remains to be established.

The proposed framework
provides a systematic
approach for investigating
the predictive contribution
of interpretable tissue
architecture alongside
pathology foundation models,
with potential applications
to more interpretable
computational histopathology.

\section*{CRediT Authorship Contribution Statement}
\textbf{Zeeshan Shahid}: Conceptualization, Methodology, Software, Investigation, Formal analysis, Visualization, Writing – original draft.\\
\textbf{Hammam M. AlGhamdi}: Conceptualization, Methodology, Supervision, Resources, Writing – review \& editing.
\section*{Funding}

This research received no external funding.

\section*{Declaration of Competing Interest}

The authors declare that they have no known competing financial
interests or personal relationships that could have appeared to
influence the work reported in this paper.

\section*{Data Availability}

The datasets analyzed in this study are publicly available.
The BCNB dataset was obtained from the Grand Challenge platform
(\url{https://bcnb.grand-challenge.org/}).
The LungHist700 dataset, described in the associated publication
in Scientific Data (2024), was obtained from Figshare
(\url{https://figshare.com/articles/dataset/25459174}).
No new clinical data were collected for this study. The analysis code is available from the corresponding authors upon reasonable request. 

\section*{Ethics Statement}

This study involved secondary analysis of publicly available histopathology datasets (BCNB and LungHist700). No new patient data or biological specimens were collected, and no direct interaction with human participants occurred. The study used the datasets in accordance with their respective access and usage conditions.
\section*{Declaration of Generative AI and AI-Assisted Technologies in the Manuscript Preparation Process}

During the preparation of this manuscript, the authors used OpenAI to assist with language refinement and Python code development. The authors reviewed and verified the resulting content and take full responsibility for the final manuscript.
	
	\bibliographystyle{plain}
	\bibliography{community.bib}
\end{document}